\pdfoutput=1

\def\year{2020}\relax
\documentclass[letterpaper]{article} 
\usepackage{aaai20}  
\usepackage{times}  
\usepackage{helvet} 
\usepackage{courier}  
\usepackage[hyphens]{url}  
\usepackage{graphicx} 
\usepackage{booktabs}
\usepackage{graphicx}  
\usepackage{pifont} 

\usepackage{tikz} 
\usepackage{comment}

\usepackage{amsmath,amsfonts,bm}

\def\eqref#1{equation~\ref{#1}}

\def\1{\bm{1}}

\DeclareMathAlphabet{\mathsfit}{\encodingdefault}{\sfdefault}{m}{sl}
\SetMathAlphabet{\mathsfit}{bold}{\encodingdefault}{\sfdefault}{bx}{n}

\usepackage{multirow}

\newcommand{\datasetName}[1]{{{{LaptopNetwork}}}{#1}}
\newcommand{\modelname}[1]{{{{Modular Supervision End-to-End Network}}}{#1}}
\newcommand{\hightlightmodelname}[1]{{{{\underline{Mo}dular \underline{S}upervi\underline{s}ion Network}}}{#1}}
\newcommand{\Modelabbreviation}[1]{{{{MOSS}}}{#1}}

\newcommand{\weixin}[1]{[{\color{blue}WX: #1}]}

\newcommand{\youzhiREF}[1]{{\textcolor{cyan}{\textit{.}}}}

\newcommand{\weixinREF}[1]{{\textcolor{yellow}{\textit{.}}}}

\definecolor{enccolor}{RGB}{141,191,227}
\definecolor{nlucolor}{RGB}{141,191,227}
\definecolor{dstcolor}{RGB}{191,161,213}
\definecolor{dplcolor}{RGB}{141,220,177}
\definecolor{nlgcolor}{RGB}{255,141,141}

\title{MOSS: End-to-End Dialog System Framework with Modular Supervision }
\newcommand{\printfnsymbol}[1]{%
  \textsuperscript{\@*}%
}

\author{
Weixin Liang\thanks{equal contribution}  \\ Computer Science Department \\ Zhejiang University \\ Hangzhou, China  \And
Youzhi Tian\printfnsymbol{1} \\ Computer Science Department \\ Zhejiang University \\ Hangzhou, China \And
Chengcai Chen \\ Xiao-i Technology\\ Shanghai, China \And
Zhou Yu \\ Computer Science Department \\  University of California, Davis \\ Davis, CA, USA
}

\begin{document}

\maketitle

\begin{abstract}

A major bottleneck in training end-to-end task-oriented dialog system is the lack of data. 
To utilize limited training data more efficiently, 
we propose \hightlightmodelname{} 
(\Modelabbreviation), an encoder-decoder training 
framework that could incorporate supervision from various intermediate dialog system modules including natural language understanding, dialog state tracking, dialog policy learning  and natural language generation. 
With only 60\% of the training data,  \Modelabbreviation-all (i.e., \Modelabbreviation{} with supervision from all four dialog modules) outperforms state-of-the-art models on CamRest676. 
Moreover, introducing modular supervision has even bigger benefits when the dialog task has a more complex dialog state and action space. 
With only 40\% of the training data, \Modelabbreviation-all outperforms the state-of-the-art model on a complex laptop network trouble shooting dataset, \datasetName{}, that we introduced. \datasetName{} consists of conversations between real customers and customer service agents in Chinese. 
Moreover, \Modelabbreviation{} framework can accommodate dialogs that have supervision from different dialog modules at both framework level and model level. Therefore, \Modelabbreviation{} is extremely flexible to update in real-world deployment.

\end{abstract}

\section{Introduction} 

\begin{figure*}[ht]
\small
\centering
\includegraphics[width=17cm]
{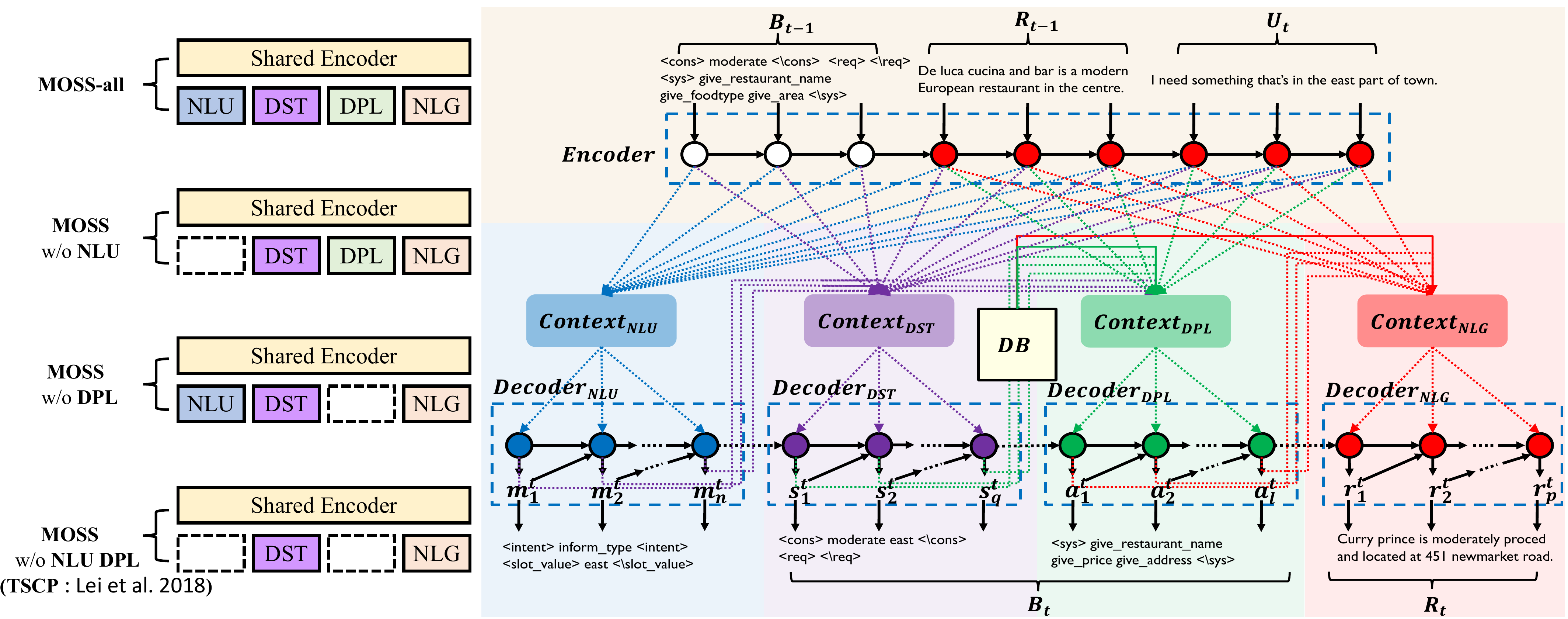}
\vspace{-0mm} 
\caption{Modular Supervision Framework (\Modelabbreviation): The left part shows 
several instances of \Modelabbreviation{} framework in plug-and-play fashion: \Modelabbreviation-all (\Modelabbreviation{} with supervision from all four dialog modules), 
\Modelabbreviation{} w/o NLU,
\Modelabbreviation{} w/o DPL,
\Modelabbreviation{} w/o NLU DPL (which is actually TSCP~\cite{lei-etal-2018-sequicity}). 
The right part shows the detailed architecture of \Modelabbreviation-all with one decoder and four decoders  (\underline{N}atural \underline{L}anguage \underline{U}nderstanding, \underline{D}ialog \underline{S}tate \underline{T}racking, \underline{D}ialog \underline{P}olicy \underline{L}earning and \underline{N}atural \underline{L}anguage \underline{G}eneration). 
The black dash lines connecting different modules represent shared hidden states. 
The colored dash lines represent modular attention, by which \Modelabbreviation-all feeds 
input to each module. 
}
\vspace{-4mm}
\label{fig:model_full}
\end{figure*}

Most current end-to-end generative dialog models require thousands of annotated dialogs to train a simple information request task \cite{lei-etal-2018-sequicity}. 
It is difficult and time consuming to collect human-human dialogs \cite{serban2015survey}. 
Due to the task constraints, it is even impossible 
to collect a large number of dialogs. 
In contrast, traditional modular framework~\cite{DBLP:journals/csl/WilliamsY07} requires less training data~\cite{DBLP:journals/dad/LowePSCLP17}. 
Traditional modular framework  
is a pipeline of the following four functional modules developed independently: 
a natural language understanding module that maps the user utterance to a distributed semantic representation;
a dialog state tracking module that accumulates the semantic representation across different turns to form the dialog state; a dialog policy learning module that decides system dialog act based on the dialog state, and a natural language generation module that maps the obtained dialog act to natural language. 
%
%
However, 
each module in such traditional modular system 
is independently optimized. Therefore,  
it is difficult to update each module whenever new training data come. 

To combine the benefits from both modular and end-to-end systems, 
we propose to follow the idea of modular systems by injecting rich supervision from each dialog module in an end-to-end trainable framework. 
Under \Modelabbreviation{} framework, 
dialog modules such as natural language understanding, dialog state tracking, dialog policy learning and natural language generation share an encoder but have their own decoders. 
Decoders of different modules are connected through hidden states rather than symbolic outputs. 
Then all the modules can be optimized jointly to avoid error propagation and model mismatch. 
In addition, since \Modelabbreviation{} produces output from individual modules during testing, 
we can easily locate the error by checking the modular output.

\Modelabbreviation{} is also a flexible framework that can be used in a plug-and-play fashion by removing  supervision from some modules. The plug-and-play feature offers options at two levels to enable full utilization of all available annotations. 
At framework level, 
for example, if the data do not have natural language understanding supervision, we can 
create a new instance (model) of \Modelabbreviation{} framework by 
removing the natural language understanding module in \Modelabbreviation. 
As a general rule of thumb, the more supervision the model has, the better the performance is, and potentially the less number of dialogs are required to reach good performance. Our results show that, MOSS-all (\Modelabbreviation{} with supervision from all four dialog modules) on only 60\% of the training data
outperforms state-of-the-art 
models on CamRest676 including TSCP~\cite{lei-etal-2018-sequicity}. 
At model level, we could patch the performance of an individual module of a specific model by adding incompletely annotated training dialogs. 
For example, 
we observe a large performance improvement of natural language generation on MOSS-all with 60\% of the training data when we add the additional 40\% training dialogs in raw format (i.e., without any annotations). 

Theoretically, %
introducing modular supervision has even bigger benefits when the dialog task has more complex dialog states and action spaces. 
To prove \Modelabbreviation's ability on complex tasks, we collect and annotate \datasetName, a dataset on the laptop network malfunction trouble-shooting task from real-world dialogs. Compared to existing datasets, \datasetName{} has a more complex and realistic dialog structure since the dialogs are between real users and professional computer maintenance engineers. Different from previous information request tasks, \datasetName{} has more actions as the dialogs are driven by the goal of fixing the network. On \datasetName, \Modelabbreviation-all (\Modelabbreviation{} with all supervision) outperforms state-of-the-art model with only 40\% of the training data.  
Based on our experiments on both \datasetName{} and CamRest676, 
we summarize the take-aways for how to efficiently build a dataset to solve a task. 
%
We will release 
the source code and the annotated \datasetName.

\section{Related work}

Different end-to-end trainable task-oriented dialog systems inject supervision differently. 
\citeauthor{kvrn} (2017) proposed to use an attention sequence-2-sequence (Seq2Seq) \cite{DBLP:journals/corr/SutskeverVL14} encoder-decoder model without intermediate dialog module's supervision except for the natural language generation part. 
Such systems require thousands of dialogs to learn one simple task.  
It is not clear if such systems can work well on complex tasks~\cite{DBLP:journals/dad/LowePSCLP17,DBLP:conf/emnlp/HeCBL18}. 
\citeauthor{DBLP:conf/sigdial/Lee14a} \citeyear{DBLP:conf/sigdial/Lee14a} 
suggested  
that there is a positive correlation between end-to-end dialog performance and dialog state tracking performance. So we believe incorporating dialog state tracking supervision will improve overall system performance.  NDM and LIDM~\cite{CamRest676,LIDM} incorporated dialog state tracking supervision via a separately-trained belief tracker. 
TSCP~\cite{lei-etal-2018-sequicity} introduced a two-decoder pipeline that combines two dialog modules together. Specifically, it jointly trained belief span decoding (dialog state tracking) and response generation. 
\citeauthor{StructuredBeliefCopyNetwork} (2018)
extended TSCP~\cite{lei-etal-2018-sequicity} 
by separately decoding information slot and predicting requested slot for dialog state tracking. 
All these approaches outperform \citeauthor{kvrn} (2017). None of them incorporated supervision from dialog policy learning. 
Though both NDM and LIDM~\cite{CamRest676,LIDM} have policy network components, they are single layer MLPs functioning as the glue that binds the system modules together. Their policy network component does not incorporate supervision from dialog policy learning. However, the dialog policy learning is important because it decides the system's next action. The system dialog act can guides the language generation. 
%
There is also work that incorporates supervision from dialog policy but not natural language understanding \cite{liu-etal-2018-dialogue}. 
However, incorporating natural language understanding supervision improve performance for tasks in which user utterances have a large number of intents and slots. 
Although \citeauthor{DBLP:conf/ijcnlp/LiCLGC17} \citeyear{DBLP:conf/ijcnlp/LiCLGC17} incorporated supervision from all four modules, it feeds the symbolic output from NLU to downstream modules and could not avoid error propagation. 
Therefore, we propose \Modelabbreviation{}, an encoder-decoder based end-to-end trainable framework that can incorporate supervision from all intermediate dialog modules, including natural language understanding (NLU), dialog state tracking (DST), dialog policy learning (DPL) and natural language generation (NLG).


Most existing task-oriented dialog datasets, such as \citeauthor{CamRest676} (2017b) and \citeauthor{MultiWOZ2} (2018), are collected in the Wizard-of-Oz (WOZ) role-play paradigm. 
In such a paradigm, the users are asked to conduct the task with detailed instruction. It improved the efficiency in collecting domain-specific data and ensures coherence and consistency between the two conversation partners. However, the user action space is relatively small compared to the real-world dialog because of the predefined constraints. 
In addition, the users are role-playing instead of having a real need to talk to the system, so the dialogs are different from practical usage. Towards tackling tasks with more dialog acts, 
\citeauthor{DBLP:journals/corr/LewisYDPB17} (2017); \citeauthor{DBLP:conf/emnlp/HeCBL18} (2018); \citeauthor{DBLP:journals/corr/abs-1906-06725} (2019) collected negotiation and persuasion dialogs by asking the two Turkers negotiate or persuade each other to reach an agreement. However, these tasks are still not collected from real users. The only real human-human real-world dialog system is a domain-specific IT helpdesk dataset \cite{DBLP:journals/corr/VinyalsL15}. But unfortunately, this dataset is not public. Therefore, to test MOSS's ability to handle complex tasks, we publish an annotated real-world dataset, \datasetName. 
It contains dialogs between real users and computer maintenance engineers on solving laptop network issues.

\section{ \Modelabbreviation: \hightlightmodelname{}}

\Modelabbreviation{} is an encoder-decoder based end-to-end trainable framework that could incorporate supervision from various intermediate dialog system modules. 
%
Figure~\ref{fig:model_full} (right) shows the detailed architecture of \Modelabbreviation-all (i.e., \Modelabbreviation{} with supervision from all four dialog modules).  
Inspired by traditional modular architecture, 
\Modelabbreviation-all has a unified encoder and four separate decoders. Each decoder aligns with a dialog module so the supervision can be introduced from each decoder. 
Between different modules, we transfer knowledge via cross-modular attention and shared hidden states without relying on symbolic outputs. 
We jointly optimize the four decoders to avoid error propagation. 
Moreover, as Figure~\ref{fig:model_full} (left) shows, 
with different instantiations, 
\Modelabbreviation{} framework can accommodate dialogs that have supervision from different dialog modules in a plug-and-play fashion.

\subsection{Methodology} 
We first present the architecture of \Modelabbreviation-all and then describe 
how the plug-and-play feature deals with incomplete annotations. 
For each dialog turn $t$, 
the system inputs are: 
the state summary of the previous turn 
$B_{t-1} = [S_{t-1}; A_{t-1}] $ 
(the concatenation of the dialog \underline{s}tate $S_{t-1}$ and the system \underline{a}ct $A_{t-1}$ of previous turn), 
the system \underline{r}esponse utterance $R_{t-1}$ of previous turn
and the \underline{u}ser utterance $U_{t}$. 
We formulate each module into a sequence-to-sequence (Seq2Seq) framework with $[B_{t-1}, R_{t-1}, U_{t}]$ as the input sequence.

\subsubsection{Natural Language Understanding (NLU) Module} 
The NLU module generates a distributed se\underline{m}antic representation $M_{t} = (m_0^t, m_1^t, \ldots, m_n^t)$ of the user utterance $U_{t}$. 
$M_{t}$ is the concatenation of user intent and the extracted values for slot filling. 
The NLU module could be formulated as: 
\begin{equation*}
M_{t} = Seq2Seq_{NLU}(B_{t-1},R_{t-1},U_{t}) 
\end{equation*}
\subsubsection{Diaog State Tracking (DST) Module} 
The DST module maintains the dialog \underline{s}tate $S_{t} = (s_0^t, s_1^t, \ldots, s_n^t)$, which is the concatenation of user expressed constraints and requests. 
DST achieves this by accumulating 
user semantic representation $M_{t}$  
across different turns $0, 1, \ldots, t$. 
So the DST module could be formulated as: 
\begin{equation*}
\begin{split}
S_{t} = Seq2Seq_{DST}(B_{t-1},R_{t-1},U_{t}|M_{t})
\end{split}
\end{equation*}
\subsubsection{Dialog Policy Learning (DPL) Module} 
The DPL module generates system act $A_{t} = (a_1^t, a_2^t, \ldots, a_l^t)$ based on the current dialog state $S_{t}$. It could be formulated as: 
\begin{equation*}
\begin{split}
A_{t} = Seq2Seq_{DPL}(B_{t-1},R_{t-1},U_{t}|M_{t}, S_{t})
\end{split}
\end{equation*}
\subsubsection{Natural Language Generation (NLG) Module} 
The natural language generation (NLG) module then maps the dialog act to 
its surface form $R_{t} = (r_1^t, r_2^t, \ldots, r_p^t) $. 
So the NLG module could be formulated as: 
\begin{equation*}
R_{t} = Seq2Seq_{NLG}(B_{t-1},R_{t-1},U_{t}|M_{t}, S_{t}, A_{t})
\end{equation*}

\subsubsection{Plug-and-Play: Dealing with Incomplete Annotations}
The plug-and-play feature offers options at both framework level and model level to deal with incomplete annotations. 
At framework level, 
to accommodate dialogs that lack supervision from different dialog modules, 
we could create different instances (models) of \Modelabbreviation{} framework 
by removing the corresponding decoder in \Modelabbreviation 
as shown in Figure~\ref{fig:model_full} (left). 
We further adopt the down-stream module(s) by removing the condition dependencies on the module(s) to be removed. 
For example, if we remove the dialog policy learning module, 
then we get \Modelabbreviation{} without supervision from dialog policy learning (\Modelabbreviation{} w/o DPL)
and re-formulate the NLG module as: 
\begin{equation*} 
R_{t} = Seq2Seq_{NLG}(B_{t-1},R_{t-1},U_{t}|M_{t}, S_{t}) 
\end{equation*} 
where $A_{t}$ is removed in the condition.

At model level, for a specific instance (model), we could patch the performance of an individual module by adding incompletely-annotated training dialogs. 
For example, if the performance of natural language generation is not satisfactory, we could add raw training dialogs without any annotations. 
For these training dialogs, we calculate the loss solely based on the natural language generation module and back-propagate the gradient to the entire model. 
The flexibility offered by these two levels of plug-and-play encourages the maximum utilization of all available annotations and the practical updates in deployed systems. 


\subsection{Building Blocks: Encoder and Decoder}
%
%

\subsubsection{Encoder} 
An encoder is shared by all modules under \Modelabbreviation{} framework. 
For each dialog turn $t$, 
a shared bidirectional GRU encodes the following three input: 
the state summary of the previous turn 
$B_{t-1} = [S_{t-1}; A_{t-1}] $ 
(the concatenation of the dialog \underline{s}tate $S_{t-1}$ and the system \underline{a}ct $A_{t-1}$ of previous turn), the system \underline{r}esponse utterance $R_{t-1}$ of previous turn and the \underline{u}ser utterance $U_{t}$. 

\begin{equation*}\label{encoder}
    \widetilde{B}_{t-1}, \widetilde{R}_{t-1}, \widetilde{U}_{t}, h_E^t = Encoder(B_{t-1},R_{t-1}, U_{t})
\end{equation*} 
where  
$\widetilde{B}_{t-1}, \widetilde{R}_{t-1}, \widetilde{U}_{t}$ 
are 
the encoder states when encoding each token of $B_{t-1}$,$R_{t-1}$ and $U_{t}$ respectively. 
$h_E^t$ is 
the last encoder hidden state.  

\subsubsection{Decoder}
The decoders in all modules (NLU, DST, DPL, NLG) have the same structure. 
Each decoder is implemented as an attention~\cite{attention} based unidirectional GRU augmented with the copy mechanism~\cite{copynet}. 
The decoder input is a sequence of distributed representations 
$ X = (x_1, x_2, \ldots, x_n)$. 
In addition, the initial decoder hidden state $h_0$ 
could 
be assigned as prior knowledge. 
The decoder output is $ Y = (y_1, y_2, \ldots, y_m)$, a sequence of the probability of output tokens. 
We also records 
$\widetilde{Y}  = (\widetilde{y}_1, \widetilde{y}_2, \ldots, \widetilde{y}_m) = (h_1, h_2, \ldots, h_m)$, 
the decoder hidden states when decoding $Y$ because it would be used by its downstream modules.  
The decoder could be formulated as:  
\begin{equation*}
    Y, \widetilde{Y} = Decoder_{\varphi}(X, h_0)
\end{equation*}
where $ \varphi $ is the module name which could be NLU, DST, DPL or NLG. The loss is defined as negative log likelihood.

%



\subsection{Natual Language Understanding (NLU) Decoder} \label{subsec:NLU}
Traditionally, a NLU module processes intent detection and slot filling separately: 
intent detection is treated as a semantic utterance classification problem, and slot filling is treated as a sequence labeling task. 
We jointly formulate intent detection and slot filling as a sequence generation problem, 
which solves multi-intent problem.

The NLU module maps user utterance $U_{t}$ to user semantic representation $M_{t}$
with the help of the information in previous turns ($B_{t-1}, R_{t-1}$).  
we formulate $Decoder_{NLU}$ as: 
\begin{equation*} 
    M_{t}, \widetilde{M}_{t} = Decoder_{NLU}([\widetilde{B}_{t-1}, \widetilde{R}_{t-1},\widetilde{U}_{t}], h_E^t)
\end{equation*} 
Note that $\widetilde{M_{t}}$ is the decoder hidden states when decoding user semantic representation $M_{t}$. 
It will be used as the input of later modules. 
The initial hidden state of $Decoder_{NLU}$ is initialized as the last hidden state $h_E^t$ of the encoder. 

\subsection{Dialog State Tracking (DST) Decoder} \label{subsec:DST} 
\Modelabbreviation{} formulates DST 
into a sequence-to-sequence framework with copy mechanism. 
So the DST module can solve the out-of-vocabulary words problem of traditional classification-based methods, as users may mention values for the informable slots which have never appeared in the training data. 
The DST module tracks dialog state $S_{t}$ by accumulating user semantic representation $M_{t}$ across different turns. 

The DST decoder also takes 
system response utterance $ \widetilde{R}_{t-1}$, user utterance $\widetilde{U}_{t}$ as input. 
Different from condensed context like state summary of previous turn $B_{t-1}$, 
$ \widetilde{R}_{t-1}, \widetilde{U}_{t}$ is the immediate dialog context of this turn. 
The immediate dialog context might contain information that's not in the condensed context. So we formulate the DST decoder as: 
\begin{equation*} 
    S_{t}, \widetilde{S}_{t} = Decoder_{DST}([\widetilde{B}_{t-1}, \widetilde{R}_{t-1}, \widetilde{U}_{t}, \widetilde{M}_{t}], \tilde{m}_{n}^t) 
\end{equation*} 
Here $Decoder_{DST}$ is initialized with the last hidden state of the NLU decoder $\tilde{m}_{n}^t$ as prior. 

\begin{figure*}[ht]
\centering
\includegraphics[width=12cm]
{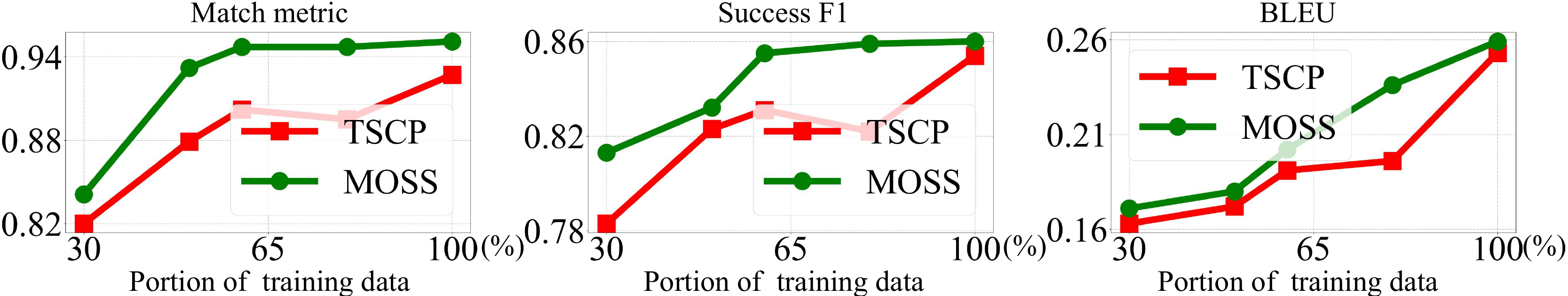}
\vspace{-2mm}
\caption{
The detailed perforamance change of TSCP and \Modelabbreviation-all on CamRest676 using a different amount of data. 
}
\vspace{-4mm}
\label{fig:DifferentPortion}
\end{figure*}

\subsection{Dialog Policy Learning (DPL) Decoder} \label{subsec:DPL}
We formulate DPL as a sequence-to-sequence problem to enable \Modelabbreviation{} to generate multiple system acts. 
The DPL module predicts the system acts $A_{t}$ by 
considering both the dialog states $S_t$ and the query results from the external database DB. 
Following~\citeauthor{snapshotLearning} (2016), 
the DPL module forms the database query 
by taking the union of the maximum values of each informable slot
in dialog state $S_{t}$ \cite{CamRest676}. 
The DB returns a one-hot vector $k_t$ 
representing different degrees of matching in the
DB (no match, 1 match, ... or more than 5 matches). 
As language model type condition~\cite{snapshotLearning}, 
$k_t$ is concatenated with the word embedding of each $a_j^t$, 
 $j \in [1,\ldots, l ] $ 
as the new embedding.  
\begin{equation*}
  emb'(a_{j}^t) = 
  \left(
  \begin{array}{c}
          emb(a_{j}^t) \\
          k_{t}
 \end{array}
 \right)
\end{equation*}
%
The DPL decoder explicitly conditions on the state summary of this turn $S_{t}$ to generate the system act $A_{t}$. 
\begin{equation*} 
    A_{t}, \widetilde{A}_{t} = Decoder_{DPL}([\widetilde{R}_{t-1}, \widetilde{U}_{t}, \widetilde{S}_{t}], \tilde{s}_{q}^t) 
\end{equation*} 
The hidden state of $Decoder_{DPL}$ is initialized as the last hidden state $\tilde{s}_{q}^t$ of $Decoder_{DST}$.

\subsection{Natural Language Generation (NLG) Decoder} \label{subsec:NLG}
The NLG decoder converts the system dialog acts $A_{t}$ into system response $R_{t}$. 
The NLG also conditions on DB query result $k_t$ in the same way as the DPL. 
The NLG decoder initializes its hidden state with the last hidden state $\tilde{a}_{l}^t$ of the $Decoder_{DPL}$ as the prior knowledge of system acts $A_{t}$. 
\begin{equation*} 
    R_{t}, \widetilde{R}_{t} = 
    Decoder_{NLG}([A_{t},\widetilde{R}_{t-1}, \widetilde{U}_{t}], \tilde{a}_{l}^t)
\end{equation*}

Finally, 
we sum up the cross-entropy losses of the four decoders and optimize the four decoders jointly 
to avoid error propagation and model mismatch: 
\begin{equation*}
\begin{split}
    & \mathcal{L}  
    = \mathcal{L}_{NLU} + \mathcal{L}_{DST} + \mathcal{L}_{DPL} + \mathcal{L}_{NLG} \\
\end{split}
\end{equation*}

\section{Restaurant Search Task}

\begin{table}[t]
\footnotesize
\begin{center}
\begin{tabular}{rlll}
\cmidrule[\heavyrulewidth]{1-4}
\multicolumn{1}{r}{\bf Model}  &\multicolumn{1}{l}{\bf Mat}  &\multicolumn{1}{l}{\bf Succ.F1} &\multicolumn{1}{l}{\bf BLEU} 
\\ 
\cmidrule{1-4}
KVRN  &N/A &N/A &0.134  \\
\cmidrule{1-4}
NDM &0.904 &0.832 &0.212 \\
LIDM &0.912 &0.840 &0.246 \\
\cmidrule{1-4}
TSCP &0.927  &0.854 &0.253\\
\cmidrule{1-4}
\Modelabbreviation{} w/o DPL & 0.932 & 0.856 & 0.251  \\
\Modelabbreviation{} w/o NLU & 0.932 & 0.857 & 0.255 \\
\cmidrule{1-4}
\Modelabbreviation-all $\times$ 60\% &0.947 &0.857 &0.202 \\
\Modelabbreviation{} $\times$ (60\%all + 40\%raw ) &0.947 & 0.859 &0.221 \\
\cmidrule{1-4}
\Modelabbreviation-all & \bf0.951 &\bf0.860 &\bf0.259 \\
\cmidrule[\heavyrulewidth]{1-4}
\end{tabular}
\caption{
Performance comparison on CamRest676 among the baselines, \Modelabbreviation-all, and several variants of \Modelabbreviation.
}
\vspace{-4mm}
\label{models_camdata}
\end{center}
\end{table}

\begin{table}[htb]
\footnotesize
\centering
\begin{tabular}{p{0.7cm}p{7cm}}
\cmidrule[\heavyrulewidth]{1-2}
\textbf{User}: & I'm looking for an expensive restaurant in the south. \\
\cmidrule{1-2}
\textbf{True}: & Hi!We have Mexican,Italian,Indian and Chinese restaurants that fit the bill.Do you have any preferences? \\
\cmidrule{1-2}
\textbf{MOSS-all}:  &  What type of food are you interested in the south part of town? (act: \textit{{\textbf{``ask\_food''}}}) \\ 
\cmidrule{1-2}
\textbf{TSCP}: & Curry Garden is a expensive Indian restaurant in the south part of town.\\
\cmidrule[\heavyrulewidth]{1-2}
\end{tabular}
\caption{An example of \Modelabbreviation-all and TSCP on CamRest676. The bold word indicates \Modelabbreviation-all's correct prediction on dialog policy.    
}
\label{table:example_camrest}
\end{table}

\begin{table}[htb]
\footnotesize
\centering
\begin{tabular}{p{8cm}}
\toprule
\textbf{Response and request}  \\
\midrule
\textbf{Sys:} Unfortunately there are no Thai restaurants in the {\textbf{north}}, do you want to change an area to look for ? \\
\textbf{User:} How about {\underline{west}} area? I also want the address, phone number, and the price range?\\ 
\midrule
\textbf{\Modelabbreviation-all.NLU:}  \textit{{\textbf{ask\_Inf}}} : {\underline{west}} address phone price \\
\midrule
\textbf{True.NLU:} \textit{{\underline{inform\_Type\_Change}}} : {\underline{west}} address phone price \\
\midrule
\textbf{\Modelabbreviation-all.DST:} constraints: Thai {\textbf{north}} requests: address phone price \\
\midrule
\textbf{True.DST:} constraints: Thai {\underline{west}} request: address phone price\\
\bottomrule
\end{tabular}
\caption{ An \Modelabbreviation-all error analysis example. The underlined words indicate the correct outputs while the bold parts indicate the incorrect outputs. 
}
\vspace{-4mm}
\label{table:example error locate}
\end{table}

We first use CamRest676 \cite{CamRest676} dataset to show \Modelabbreviation's advantage on existing task-oriented dialog datasets.
We annotate CamRest676 with five user intents 
(e.g., \textit{Inform\_type\_change}, \textit{Goodbye}) 
and 10 system dialog acts 
(e.g., \textit{give\_foodtype}, \textit{ask\_food}). 
We follow \citeauthor{CamRest676} (2017b); \citeauthor{lei-etal-2018-sequicity} (2018) to split the data as 3:1:1 for training, validation and testing.

\subsection{Baselines and Metrics}\label{sec:cammetrics}
We compare our model against a set of state-of-the-art models: 
\textit{(\romannumeral1)}
KVRN \cite{kvrn} is an attention seq2seq encoder-decoder model without intermediate dialog module's supervision except for the natural language generation;   
\textit{(\romannumeral2)} 
NDM \cite{CamRest676} and 
\textit{(\romannumeral3)} 
LIDM \cite{LIDM} incorporate dialog state tracking supervision via a separately-trained belief tracker; 
\textit{(\romannumeral4)} 
TSCP \cite{lei-etal-2018-sequicity} could be viewed as an instance of \Modelabbreviation{} without supervision from natural language understanding and dialog policy learning. 
\textit{(\romannumeral5 - \romannumeral8)} 
We also evaluate some variants of MOSS shown in Figure~\ref{fig:model_full} (left). 
Following~\citeauthor{lei-etal-2018-sequicity} (2018), we use three evaluation metrics: entity match rate (Mat) on dialog state, success F1 (Succ.F1) on requested slots and BLEU \cite{DBLP:conf/acl/PapineniRWZ02} on generated system utterances.

\begin{figure*}[ht]
\centering
\includegraphics[width=16cm]{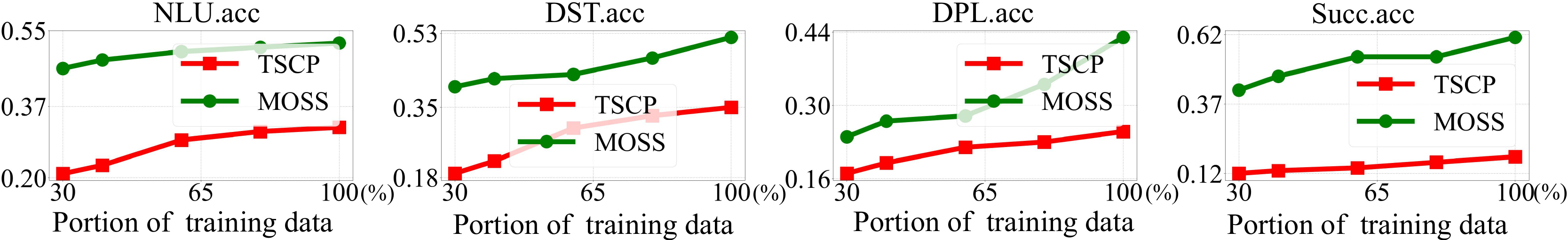}
\vspace{-2mm}
\caption{
The detailed performance change of TSCP and \Modelabbreviation-all on \datasetName{}  using a different amount of data. 
}
\vspace{-3mm}
\label{fig:DifferentPortion2}
\end{figure*}

\subsection{Results}\label{CamRestResults}

The first key takeaway is that the more supervision the model has, the better the performance is. 
As shown in Table~\ref{models_camdata}, 
in terms of overall performance, 
we have 
\textit{(\romannumeral1)} KVRN $<$ 
\textit{(\romannumeral2)} NDM $\approx$ 
\textit{(\romannumeral3)} LIDM $<$ 
\textit{(\romannumeral4)} TSCP $<$ 
\textit{(\romannumeral5)} MOSS w/o DPL $\approx$ 
\textit{(\romannumeral6)} MOSS w/o NLU $<$ 
\textit{(\romannumeral9)} MOSS-all. 
We note that this performance ranking is the same as the ranking of how much supervision each system receives: 
\textit{(\romannumeral1)} KVRN only incorporates supervision from one dialog module (i.e., natural language generation); 
\textit{(\romannumeral2, \romannumeral3, \romannumeral4)} NDM, LIDM, TSCP incorporate supervision from 
two dialog modules (i.e., dialog state tracking and natural language generation); 
\textit{(\romannumeral5, \romannumeral6)} 
MOSS without dialog policy learning  (MOSS w/o DPL) and 
MOSS without natural language understanding (MOSS w/o NLU) 
incorporate supervision from three dialog modules; 
\textit{(\romannumeral9)} MOSS-all incorporates supervision from all four modules and outperform all models on all three metrics.


Another takeaway is that models that have access to more detailed supervision need fewer number of dialogs to reach good performance. 
Row 7 in Table~\ref{models_camdata} shows that 
with only 60\% training data, 
\Modelabbreviation-all outperforms state-of-the-art baselines in terms of task completion (Mat and success F1). 
As for language generation quality (BLEU), 
\Modelabbreviation-all with 60\% training data performs worse. 
We suspect that it is partially because \Modelabbreviation-all with 60\% training data has seen fewer number of dialogs and thus has a weaker natural language generation module. We validate this hypothesis by training \Modelabbreviation-all with 60\% training data with all annotations plus the left 40\% training data without any annotation (i.e., \Modelabbreviation-all $\times$ 60\% + 40\%raw, Row 8 in Table~\ref{models_camdata}). We observe a large improvement on the BLEU score.

\Modelabbreviation-all $\times$ 60\% + 40\%raw also shows the plug-and-play feature at model level. 
An instance of \Modelabbreviation{} framework (e.g., \Modelabbreviation-all) 
could accommodate dialogs that have supervision from different dialog modules (e.g., all four modules v.s. only natural language generation module). 
The plug-and-play feature at model level allows us to 
patch the performance of an individual module (e.g., natural language generation) by adding incompletely annotated dialogs.

Compared to \Modelabbreviation-all with only 60\% training data, 
\Modelabbreviation-all using all data only improves the performance slightly from 0.947 to 0.951 in Mat and 0.857 to 0.867 in Succ.F1. 
The improvement is not huge because restaurant search is a relatively simple task. 
TSCP's performance drops drastically by reducing the training data (from 0.927 to 0.902 in Mat and 0.854 to 0.831 in Succ.F1). 
If limited training data is available, MOSS would potentially outperform TSCP much more significantly (0.947 VS 0.902 in Mat and 0.857 VS 0.831 in Succ.F1). Figure~\ref{fig:DifferentPortion} shows the detailed performance change between the two models using a different amount of data. 


\subsubsection{Case Study}
Since TSCP is the best among all the baselines, we select TSCP to compare against MOSS in the case study. 
Table~\ref{table:example_camrest} presents an example from the testing set. We found after incorporating supervision from dialog policy MOSS performs better than TSCP. 
MOSS-all learns to ask the user for more information (act: {``ask\_food''}) 
when there are too many matched results in the database. 
In contrast, TSCP instead acts as there is only one restaurant satisfying the user's constraint, 
though TSCP tracts the dialog state correctly. 
We suspect this error is caused because TSCP replies with the utterance it has seen the most in a similar context without distinguishing even similar context may lead to different dialog act choice.

\subsubsection{Error analysis}
The output from individual modules in \Modelabbreviation{} helps to locate its error easily. 
Table~\ref{table:example error locate} shows an error in the generated dialog state of \Modelabbreviation{}  
({``north''} v.s. {``west''}). The natural language understanding produced correct slots 
but in the dialog act intent prediction 
(\textit{{``ask\_info''}} v.s.  \textit{{``inform\_type\_change''}}), it produced wrong values. 
So the DST receives the wrong information. For such errors, given that \textit{{``inform\_type\_change''}} occurs much less than other tags 
like \textit{{``ask\_info''}}, one solution is to collect more examples on these two confusing dialog acts for training.

\section{{Laptop Network Troubleshooting}}


In this section, we first introduce a complex laptop network troubleshooting dataset-\datasetName. We then evaluate MOSS on \datasetName, showing that when the dialog task has a more complex dialog state and action space, introducing modular supervision has even bigger benefits.

\subsection{\datasetName{} Dataset} \label{subsec:dataset}

We collect \datasetName{}, a real-world laptop network troubleshooting task in Chinese. 
Different from dialogs generated by crowd-source workers \cite{CamRest676}, 
\datasetName{} is more realistic since it involves real customers with technical problems and professional computer maintenance engineers on an online typing after-sales service platform. 
In \datasetName{}, the engineer might give different solutions based on the feedback from customers. 
\datasetName{} contains 618 dialogs from 97 engineers with 14.76 turns per dialog, 51.67 Chinese characters per utterance and 985.82 seconds of time elapsed per dialog on average. 

Developing an annotation scheme for 
\datasetName{} is challenging because 
dialogs are full of computer maintenance terminologies such as ``network adaptor'' and ``winsock'' that are difficult to understand for naive annotators. So to guarantee the annotation quality, 
we hired two well-trained computer technicians and went through many iterations to polish the dialog annotation scheme. Finally we identified 
60 different dialog acts with 12 of them as network solution related dialog acts. Detailed annotation is presented in Appendix. 
Compared to dataset on information search task like CamRest676 which only has two dialog acts: request and provide information, 
our proposed network troubleshooting task is much more complex (Figure~\ref{fig:Actdatasetcompare} in Appendix~\ref{Appendix:dataset}) 
as the system needs to first understand users' problem and then collaborate with user to perform troubleshooting to resolve the problem. After finalizing the dialog annotation scheme, 
the two technicians then annotate each utterance with one or more dialog acts. 
The two experts reached  $ \kappa = 0.852 $  inter-annotator agreement on 300 turns.

\begin{figure*}[ht]
\centering
\includegraphics[width=12cm]
{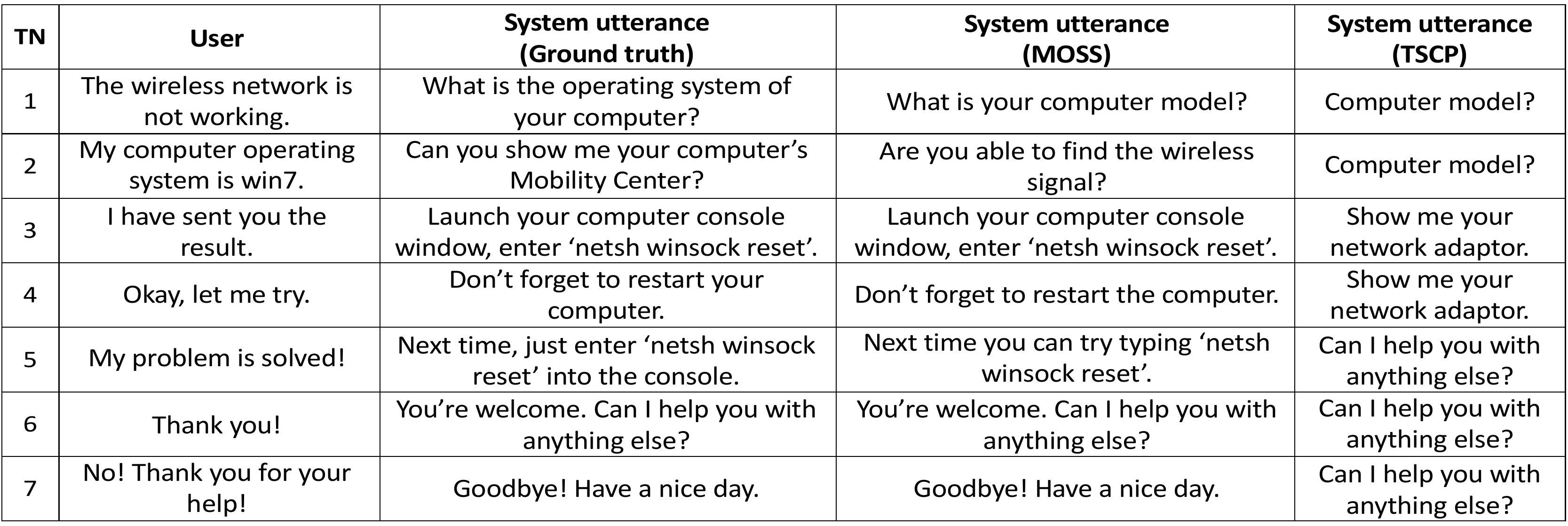}
\caption{An example dialog generated by \Modelabbreviation-all and TSCP. TN denotes the turn number. }
\vspace{-4mm}
\label{table:example network}
\end{figure*}

\begin{table}[ht]
\footnotesize
\begin{center}
\begin{tabular}{rlllll}
\cmidrule[\heavyrulewidth]{1-6}
\multicolumn{1}{r}{\bf \scriptsize Model}  &\multicolumn{1}{l}{\bf \scriptsize NLU.acc} &\multicolumn{1}{l}{\bf \scriptsize DST.acc} &\multicolumn{1}{l}{\bf \scriptsize DPL.acc} &\multicolumn{1}{l}{\bf \scriptsize Succ.acc}  &\multicolumn{1}{l}{\bf \scriptsize BLEU} \\ 
\cmidrule{1-6}
{\scriptsize TSCP} &0.32 &0.35 &0.25  &0.18 &  0.050 \\  
\cmidrule{1-6}
{\scriptsize \Modelabbreviation{} w/o DPL} & 0.51 & 0.34  & -- & -- & 0.109 \\ 
{\scriptsize \Modelabbreviation{} w/o NLU} & -- & 0.45  & 0.40 & 0.50 & 0.115 \\ 
\cmidrule{1-6}
{\scriptsize \Modelabbreviation{} $\times$ 40\%} &0.48 &0.42 &0.27 &0.47 &0.063  \\
\cmidrule{1-6}
{\scriptsize \Modelabbreviation-all} &\bf0.52 &\bf0.52 &\bf0.43  &\bf0.61 &\bf0.122 \\ 
\cmidrule[\heavyrulewidth]{1-6}
\end{tabular}
\caption{
Performance comparison on \datasetName{} among 
\Modelabbreviation-all, TSCP, 
and several variants of \Modelabbreviation. 
}
\vspace{-6mm}
\label{models_ourdata1}
\end{center}
\end{table}

\subsection{Baselines and Metrics}

Table~\ref{models_camdata} shows TSCP  \cite{lei-etal-2018-sequicity} perform the best among the baselines on CamRest676. So we compare our model against  TSCP and 
some variants of MOSS 
on \datasetName{}. 
We augment the belief span $B_t$ originally introduced in \citeauthor{lei-etal-2018-sequicity} (2018) by concatenating user act, old $B_t$ and system act in TSCP. 
This augmentation makes sure that TSCP has access to the same annotations as \Modelabbreviation, 
otherwise TSCP could hardly generate reasonable response.

Since \datasetName{} is more complex than CamRest676, we add more metrics to capture different perspectives for model performance evaluation. 
To evaluate the performance of all four modules respectively, 
we calculate: 
natural language understanding accuracy \textbf{NLU.acc}, the accuracy of user dialog act and slots; 
dialog state tracking accuracy \textbf{DST.acc}, the accuracy of user expressed constraints and requests; 
dialog policy learning accuracy \textbf{DPL.acc}, the accuracy of system dialog act and slots. 
In \datasetName{}, whether the system can give an accurate solution to solve the problem is important. So we design \textbf{Succ.acc} to capture the system's task completion rate.  
Because the task is very complex, as long as the system provides the correct solution, the task is considered successful.

\subsection{Results}

As expected, introducing modular supervision has even bigger benefits when the dialog task has a more complex dialog state and action spaces. 
As shown in Table~\ref{models_ourdata1}, 
with only 40\% training data, \Modelabbreviation-all can outperform the TSCP on all the metrics. 
Figure~\ref{fig:DifferentPortion2} shows a consistent large performance gap between TSCP and MOSS-all on \datasetName{} using a different amount of data.

With 100\% training data, 
\Modelabbreviation-all significantly outperforms TSCP on all the metrics mentioned above. For task completion rate (Succ.acc), \Modelabbreviation-all outperforms the state-of-the-art model by \textbf{42\%}. We suspect that the big performance boost comes from the additional modular supervision \Modelabbreviation-all has. For the complex task, user dialog act and system dialog act are very effective supervision to facilitate dialog system learning. 
Without such supervision, 
we observe that TSCP tends to repeat trivial system responses that are frequently seen in the training data (more details in Case Study). 
Therefore, TSCP achieves moderate, but not high scores for all the metrics. 
\Modelabbreviation-all also outperforms the state-of-the-art model by \textbf{7\%} in language generation quality. It is not surprising that with the supervision from DPL, the generated dialog act can guide the NLG module to generate a response with the correct intent.


We now examine the performance change in each perspective when removing 
dialog policy learning module (MOSS w/o DPL) or natural language understanding module (MOSS w/o NLU). 
Without dialog policy learning module, 
MOSS w/o DPL achieves comparable natural language understanding accuracy (NLU.acc)  but degraded dialog state tracking accuracy (DST.acc) and natural language generation quality (BLEU). 
Without dialog policy learning module, 
MOSS w/o DPL exhibits difficulty in directly learning the correlation between  dialog state tracking and natural language generation. 
Without natural language understanding module, 
MOSS w/o NLU 
lacks the semantic information from user utterance and 
performs worse in downstream tasks (i.e., dialog state tracking, dialog policy learning, natural language generation).

\subsubsection{Case Study}
Figure~\ref{table:example network} presents an example in \datasetName. 
Without supervision from NLU and DPL, it is difficult 
to generate correct system acts and responses in complex tasks. So TSCP tends to repeat trivial system responses (turn 1\&2 ; turn 3\&4; turn 5\&6\&7) that are frequently seen in the training data. 
In contrast, 
with supervision from NLU and DPL, 
\Modelabbreviation{} understands the dialog context better and reacts with proper system acts and responses:  
\Modelabbreviation{} is able to make inquiries (turn 1\&2), give solutions (turn 3), remind users important steps in the solution (turn 4) 
and close the dialog politely (turn 5\&6\&7). 

\section{Discussion} 





Our experiments provide some guidance for managing the budget of constructing a new dialog dataset. 
For dialog tasks that have more complex dialog states and action space like \datasetName, 
supervision from all four modules leads to much higher performance and requires significantly fewer number of dialogs (e.g., 40\% in \datasetName). Therefore, annotating natural language understanding and dialog policy learning should be prioritized during the construction of such datasets. For simple dialog tasks like information search tasks (e.g., CamRest676), the benefits of adding more supervision is still huge. Moreover, it is possible to automatically annotate the natural language understanding and dialog policy learning in these simple tasks. In CamRest676 for example, we obtain annotations for natural language understanding by calculating the difference of the current and previous dialog states. We also obtain annotations for dialog policy learning by reusing the regular expressions designed for delexicalization of system response in ~\cite{CamRest676}. Although collecting more dialogs is important, if it is possible to get detailed annotations for free, we suggest to incorporate these supervision first.




\section{Conclusion}

We propose \hightlightmodelname{} 
(\Modelabbreviation), an end-to-end trainable framework that incorporates supervision from various intermediate dialog system modules. 
Our experiments show that the more supervision the model has, the better the performance. If more supervision is included, the model needs less number of training dialogs to reach state-of-the-art performance.
In addition, such benefit is observed even larger when the dialog task has a more complex dialog state and action space for example, \datasetName. We  introduce \datasetName, which is a complex real-world laptop network malfunction trouble-shooting task.
Moreover, MOSS framework accommodates dialogs that have supervision from different dialog modules at both framework level and model level. 
At framework level we create different models with different modules removed; at model level we support feeding dialogs with annotations for different modules into the same model. Such property is extremely useful in real-world industry setting.  
\bibliography{ref}
\bibliographystyle{aaai}
\clearpage
\section{Dataset Supplemental Material}\label{Appendix:dataset}

\begin{figure}[ht]
\flushright
\includegraphics[width=7cm]
{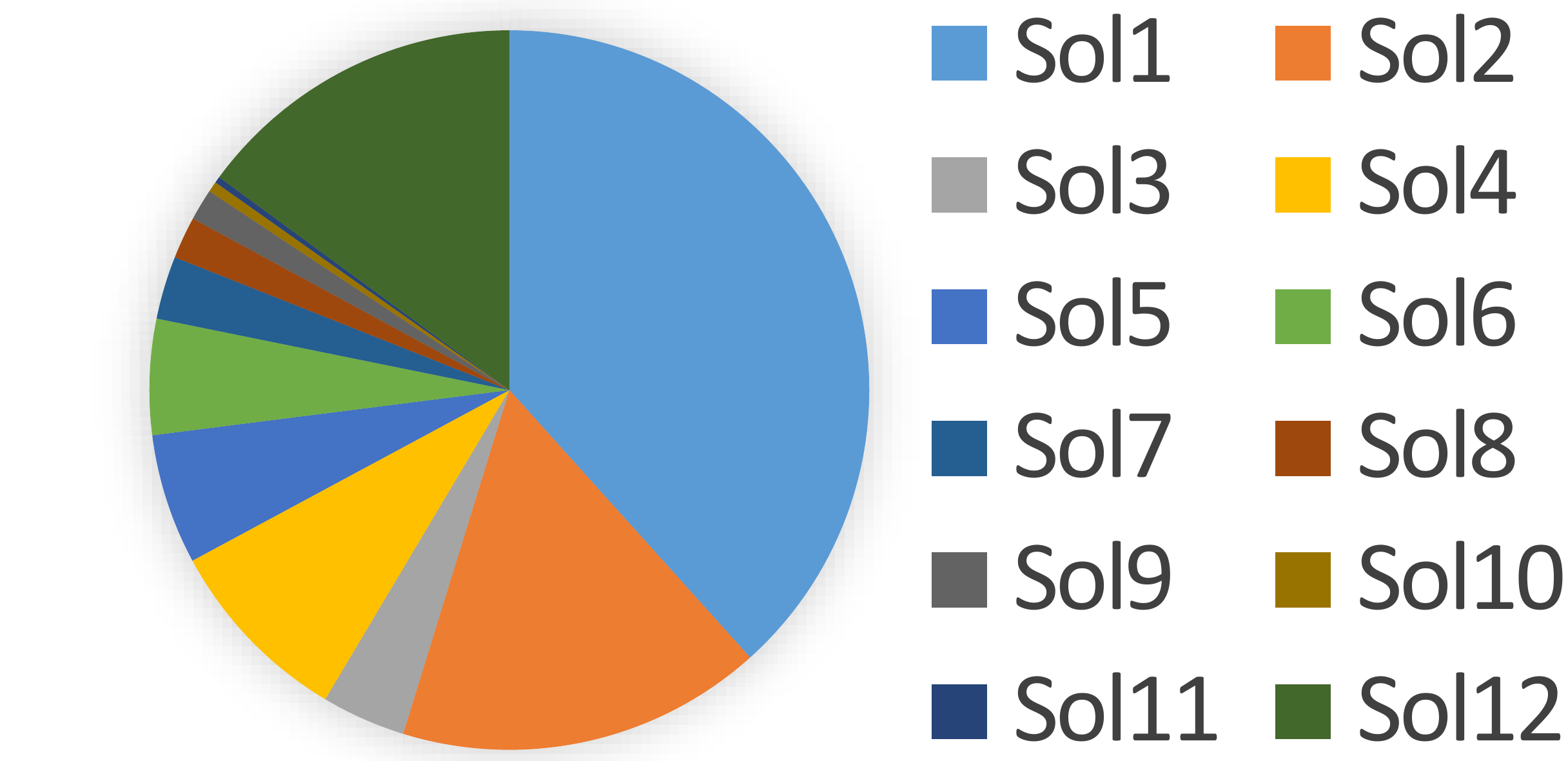}
\vspace{-0mm}
\caption{The distribution of 12 network solution related dialog acts. The reason of such a screw distribution is that \datasetName{} is collected from real world. 
}
\vspace{-4mm}
\label{fig:solutionPortion}
\end{figure}

\begin{figure}[ht]
\centering
\includegraphics[width=5cm]
{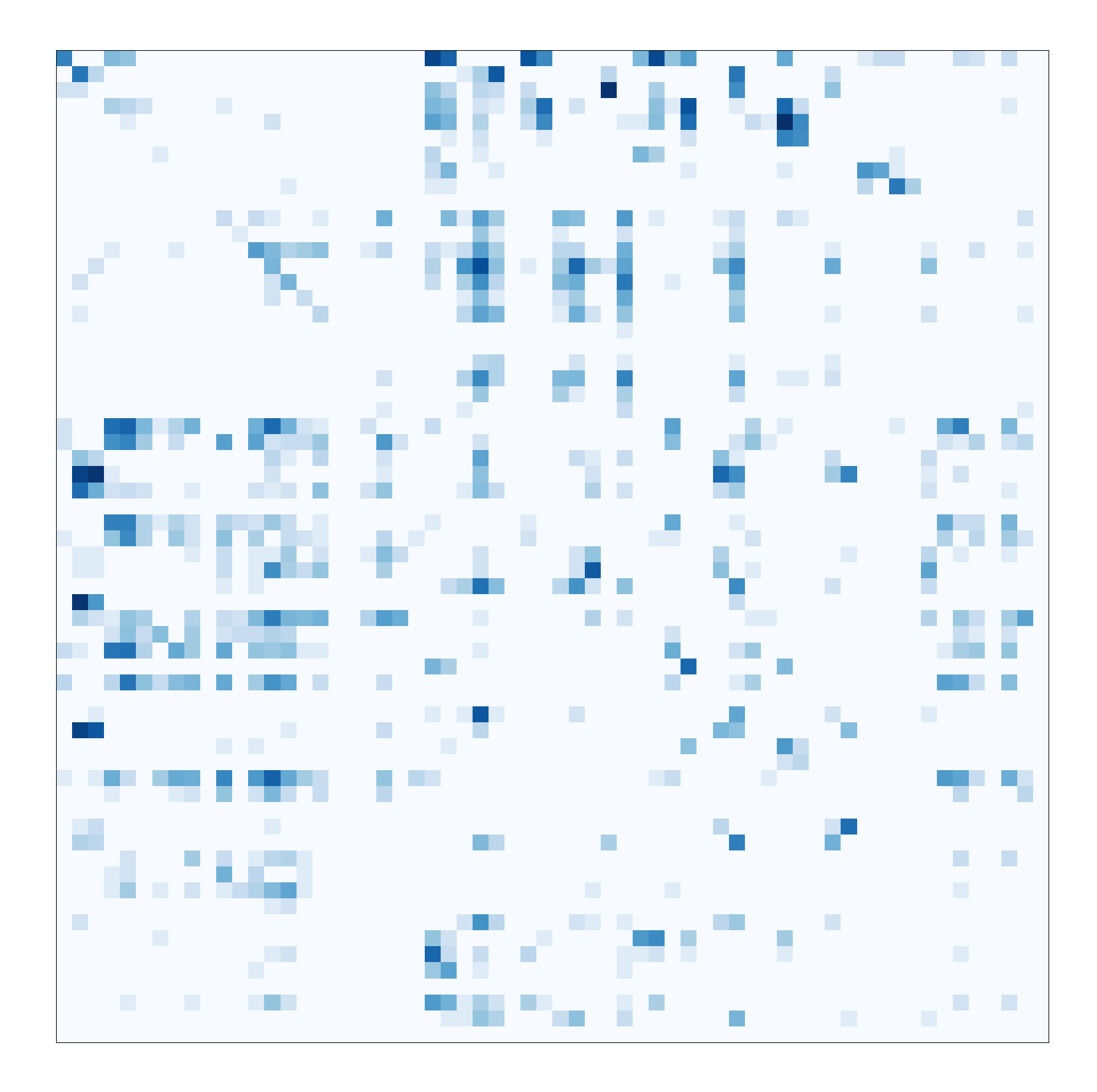}
\vspace{-0mm}
\caption{The bi-gram transition probability matrix of the raw tags.
}
\vspace{-4mm}
\label{fig:transitionProbMat}
\end{figure}


\begin{table}[ht]
\small
\centering
\begin{tabular}{rl}
\toprule
Statistics & \datasetName{} \\
\midrule
Total dialogs  & 618 \\
\midrule
Number of Engineer & 93 \\
\midrule
Number of Customer & 618 \\ 
\midrule
Average Speaker Turns Per dialog & 14.76 /2.0  \\
\midrule
Average Tokens Per dialog       & 814.31 \\
\midrule 
Average Tokens Per Utterance      & 51.67 \\
\midrule
Average Time (Seconds) Per Dialog  & 985.82 \\
\bottomrule
\end{tabular}
\caption{Basic Statistics of the \datasetName{}.
}\label{table:RawDatasetStatistics}
\end{table}

\begin{figure}[t]
\centering
\includegraphics[width=7cm]
{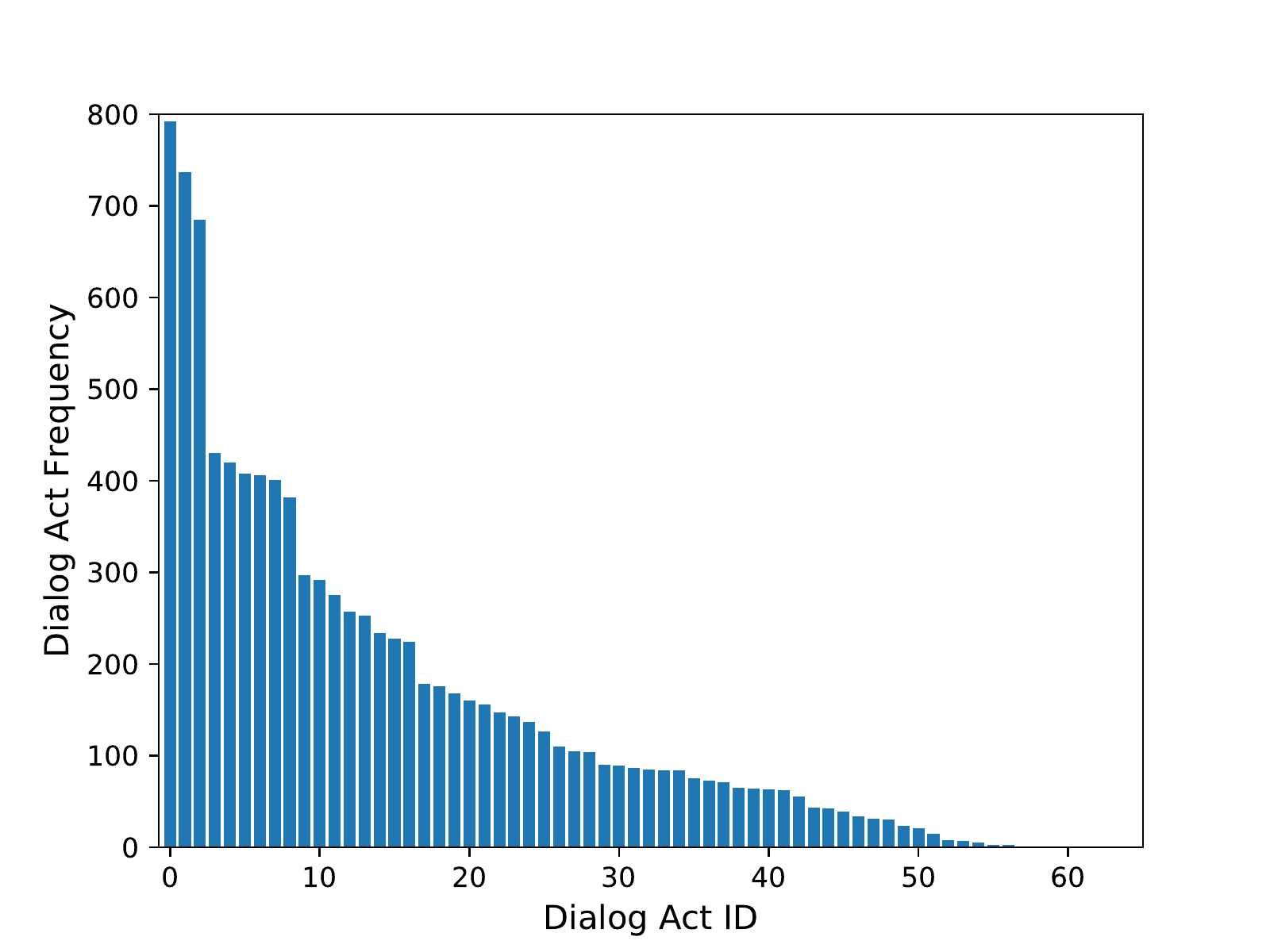}
\vspace{-0mm}
\caption{A plot of dialogue acts frequency in \datasetName{}. 
Such a long-tail distribution of dialog acts 
indicate \datasetName{} has a complex action space. 
Tag IDs are assigned based on tag frequency. For example, tag ID 1-4 correspond to four most common tags, i.e., “(system) greeting”, “(system) goodbye”, “(user) task completed” and “(system) what else can help”. 
}
\vspace{-5.0mm}
\label{fig:log_log}
\end{figure}

\subsection{More Statistics of \datasetName{}}
We first give the basic statistics of \datasetName{} in Table~\ref{table:RawDatasetStatistics}. 
\datasetName{} contains dialogs with 14.76 turns per dialog, 51.67 Chinese characters per utterance and 985.82 seconds of time elapsed per dialog on average. These numbers shows that our dataset is challenging to be learned by dialog systems. 
To explore \datasetName{}, 
we propose a dialog annotation scheme. There are 60 different dialog acts and 12 of them are network solution related dialog acts. 
In contrast, CamRest676 only has two dialog acts: request and provide information. Even if we extend such two acts with their associated slots to generate more fine-grained acts, there are only 9 acts, which is much fewer than us. 
Figure~\ref{fig:Actdatasetcompare} presents all dialog acts in \datasetName and compare them against those in CamRest676. 
The reason why \datasetName{} has more actions 
compared to previous information request tasks like CamRest676 
is that 
the dialogs are driven by the goal of fixing the network. 
Figure~\ref{fig:solutionPortion} presents the portion of each solution in \datasetName. 
Figure~\ref{fig:transitionProbMat} presents the transition probability matrix of the raw tags. This shows the complexity of the action space in \datasetName. 
All these characteristics indicate \datasetName{} has a complex dialog structure because these dialogs are generated by real-world users. 
%

\subsection{Dataset Comparison}

\begin{figure*}[htb]
\centering
\includegraphics[width=14cm]{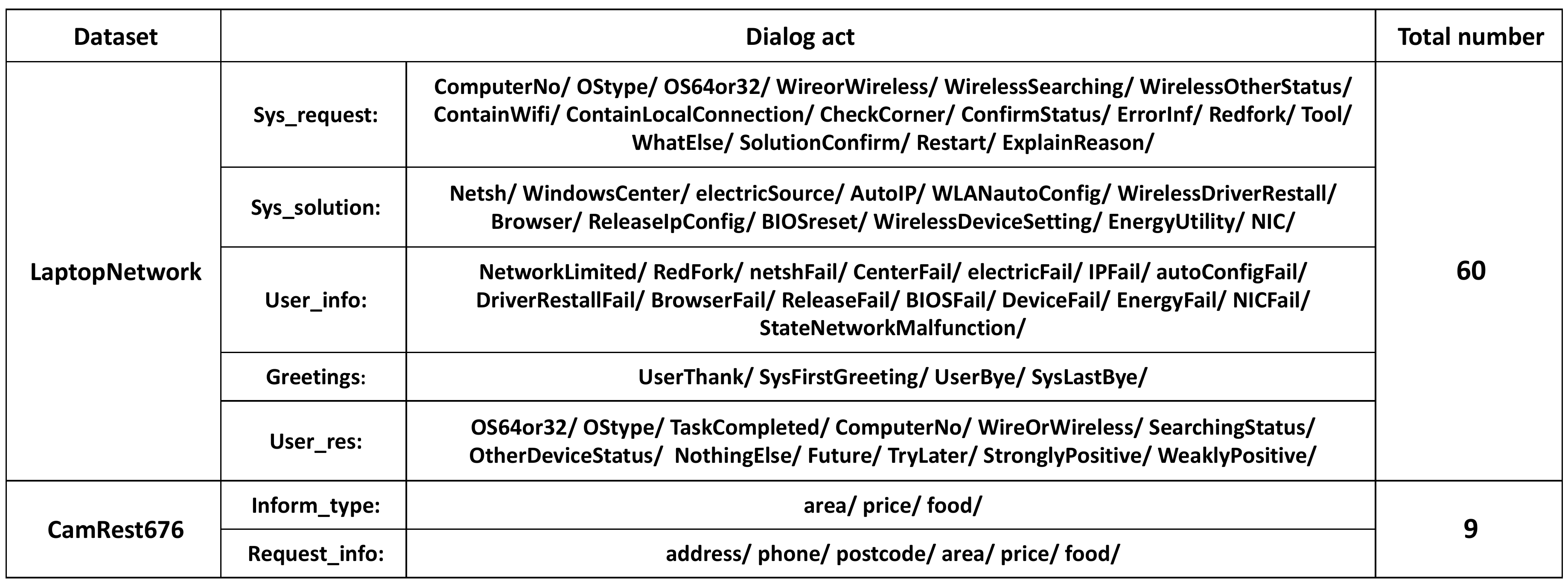}
\vspace{-0mm}
\caption{The comparison of statics information between \datasetName{} and CamRest676}
\label{fig:Actdatasetcompare}
\end{figure*}

Here we present the comparison between \datasetName{} and CamRest676 as shown in Figure~\ref{fig:Actdatasetcompare}. CamRest676 only have 9 different types of the dialog act and slot. In contrast, \Modelabbreviation{} has a total number of 60 different dialog acts. 



\section{Parameters settings}\label{Appendix:parameters setting}

For both CamRest676 and \datasetName{}, we use the same parameters settings. The vocabulary size is 800. The embedding size is 50. The hidden state size for both encoder GRU and decoder GRU is 50. The learning rate $lr$ is 0.003. After 10 epochs, the learning rate will decay as a rate of 0.5. The batch size is 32. The drop rate is 0.5. The max training epoch is 11. 


\section{Example}\label{Appendix:Chinese example}

\begin{figure*}[htb]
\centering
\includegraphics[width=16cm]{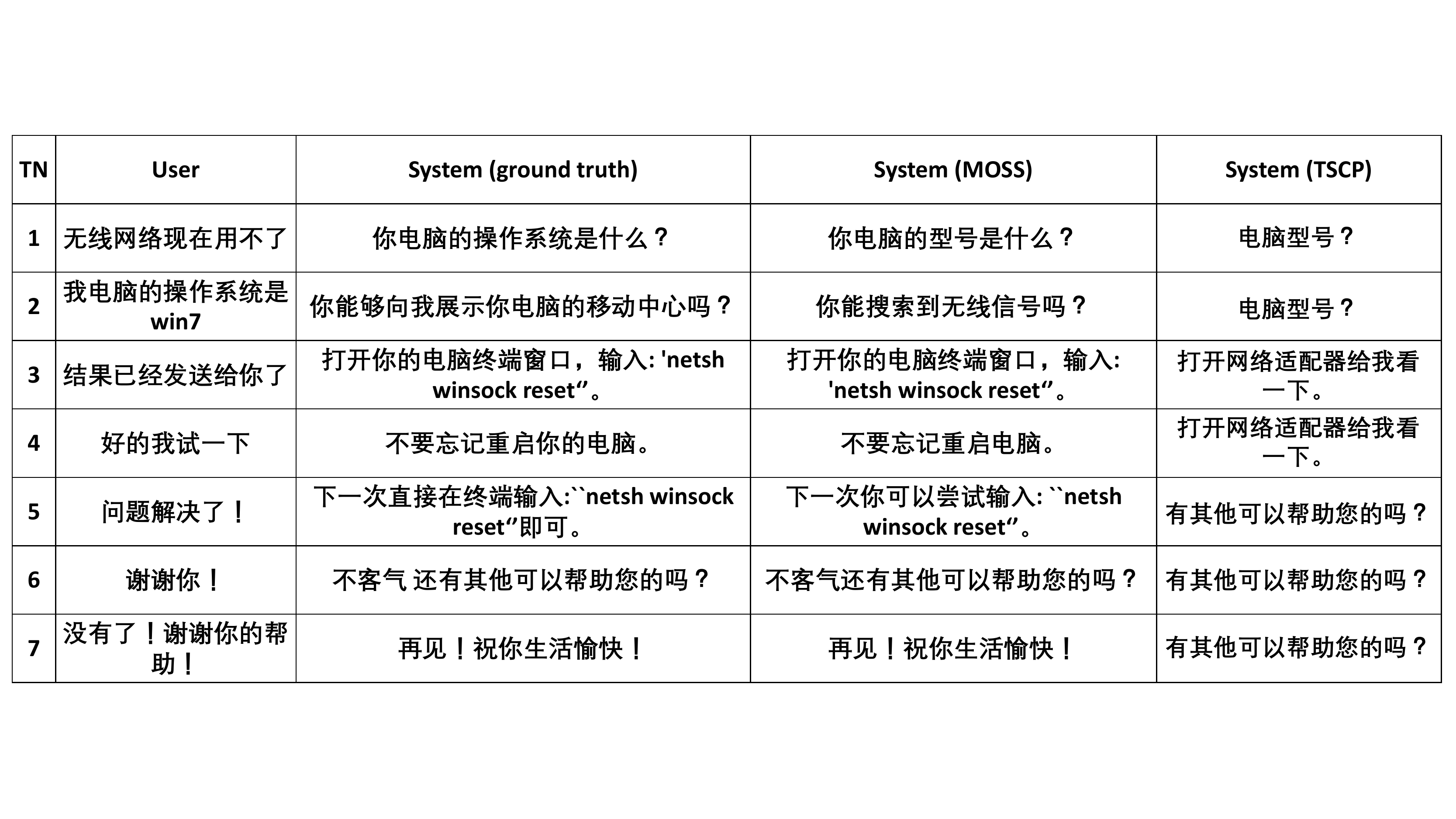}
\vspace{-0mm}
\caption{Dialog example in \datasetName{} in Chinese.}
\vspace{-4mm}
\label{table:Chinese example network}
\end{figure*}

Figure~\ref{table:Chinese example network} shows the Chinese version of the example in Table~\ref{table:example network}. The first column show the turn number. The second column shows the user utterance and the third column shows the ground truth system utterance. The last two columns show the generated system utterance by \Modelabbreviation{} and TSCP.


\end{document}